\pdfoutput=1

\documentclass[11pt]{article}

\usepackage{acl}

\usepackage{times}
\usepackage{latexsym}

\usepackage[T1]{fontenc}

\usepackage[utf8]{inputenc}

\usepackage{microtype}

\usepackage{amsmath}
\usepackage{amssymb}
\usepackage{xcolor}
\usepackage{arydshln}
\usepackage{tikz}
\usetikzlibrary{arrows, decorations.text, shapes.geometric, positioning, decorations.pathreplacing, calligraphy}
\usepackage{pgfplots}
\usepackage{pgfmath}
\usepackage{pgffor}
\pgfplotsset{compat=1.17}
\usepackage{subcaption}
\usepackage{mathtools}
\usepackage{multirow}
\usepackage{booktabs}
\usepackage{pifont}
\usepackage{ifthen}

\newcommand{\cref}{\colorbox{blue!10}{\textsc{Ref}}}
\newcommand{\csrc}{\colorbox{red!10}{\textsc{Src}}}
\newcommand{\csrcref}{\colorbox{green!10}{\textsc{Src+Ref}}}
\newcommand{\ccref}{\fboxsep0.5ex\colorbox{blue!10}{\textsc{Ref}}}
\newcommand{\ccsrc}{\fboxsep0.5ex\colorbox{red!10}{\textsc{Src}}}
\newcommand{\ccsrcref}{\fboxsep0.5ex\colorbox{green!10}{\textsc{Src+Ref}}}

\newcommand{\cmodel}{UniTE}

\title{\cmodel: Unified Translation Evaluation}

\author{Yu Wan$^{a,b}$\thanks{~~Work was done when Yu Wan was interning at DAMO Academy, Alibaba Group.}~~~Dayiheng Liu$^b$\thanks{~~Dayiheng Liu and Derek F. Wong are co-corresponding authors.}~~~Baosong Yang$^b$~~~Haibo Zhang$^b$~~~Boxing Chen$^b$\\\textbf{Derek F. Wong}$^{b\dagger}$~~~\textbf{Lidia S. Chao}$^a$ \\
  $^a$NLP$^2$CT Lab,
  University of Macau\\
  {\tt nlp2ct.ywan@gmail.com, \{derekfw,lidiasc\}@umac.mo} \\
  $^b$Alibaba Group\\
  {\tt \{liudayiheng.ldyh,yangbaosong.ybs,zhanhui.zhb,}\\
  {\tt boxing.cbx\}@alibaba-inc.com}}

\begin{document}
\maketitle
\begin{abstract}
Translation quality evaluation plays a crucial role in machine translation. According to the input format, it is mainly separated into three tasks, \textit{i.e.}, reference-only, source-only and source-reference-combined.
Recent methods, despite their promising results, are specifically designed and optimized on one of them.  
This limits the convenience of these methods, and overlooks the commonalities among tasks.
In this paper, we propose \cmodel, which is the first unified framework engaged with abilities to handle all three evaluation tasks.
Concretely, we propose monotonic regional attention to control the interaction among input segments, and unified pretraining to better adapt multi-task learning. 
We testify our framework on WMT 2019 Metrics and WMT 2020 Quality Estimation benchmarks. Extensive analyses show that our \textit{single model} can universally surpass various state-of-the-art or winner methods across tasks.
Both source code and associated models are available at \href{https://github.com/NLP2CT/UniTE}{https://github.com/NLP2CT/UniTE}.

\end{abstract}

\section{Introduction}
Automatically evaluating the translation quality with the given reference segment(s), is of vital importance to identify the performance of Machine Translation (MT) models~\cite{freitag2020bleu,mathur2020tangled,zhao2020limitations,kocmi2021ship}.
Based on the input contexts, translation evaluation can be mainly categorized into three classes:
1) reference-only evaluation (\cref) approaches like BLEU~\cite{papineni2002bleu} and BLEURT~\cite{sellam2020bleurt}, which evaluate the hypothesis by referring the golden reference at target side;
2) source-only evaluation (\csrc) methods like YiSi-2~\cite{lo2019yisi} and TransQuest~\cite{ranasinghe2020transquest}, which are also referred as quality estimation (QE).
These methods estimate the quality of the hypothesis based on the source sentence without using references;
3) source-reference-combined evaluation (\csrcref) works like COMET~\cite{rei2020comet}, where the evaluation exploits information from both source and reference.
With the help of powerful pretrained language models~\citep[PLMs,][]{devlin2019bert,conneau2020unsupervised}, model-based approaches (\textit{e.g.}, BLEURT, TransQuest, and COMET) have shown promising results in recent WMT competitions~\cite{ma2019results,mathur2020results,freitag2021results,fonseca2019findings,specia2020findings,specia2021findings}.

Nevertheless, each existing MT evaluation work is usually  designed for one specific task, \textit{e.g.}, BLEURT is only used for \cref~task and can not support \csrc~and \csrcref~tasks.
Moreover, those approaches preserve the same core -- evaluating the quality of translation by referring to the given segments.
We believe that it is valuable, as well as feasible, to unify the capabilities of all MT evaluation tasks (\cref, \csrc~and \csrcref) into one model.
Among the promising advantages are ease of use and improved robustness through knowledge transfer across evaluation tasks.
To achieve this, two important challenges need to be addressed:
1) How to design a model framework that can unify all translation evaluation tasks?
2) How to make the powerful PLMs better adapt to the unified evaluation model?

In this paper, we propose \textbf{\cmodel} - \textbf{Uni}fied \textbf{T}ranslation \textbf{E}valuation, a novel approach which unifies the functionalities of \cref, \csrc~and \csrcref~tasks into one model.
To solve the first challenge as mentioned above, based on the multilingual PLM, we utilize layerwise coordination which concatenates all input segments into one sequence as the unified input form.
To further unify the modeling of three evaluation tasks, we propose a novel Monotonic Regional Attention (MRA) strategy, which allows partial semantic flows for a specific evaluation task. For the second challenge, a multi-task learning-based unified pretraining is proposed.
To be concrete, we collect the high-quality translations and degrade low-quality translations of NMT models as synthetic data.
Then we propose a novel ranking-based data labeling strategy to provide the training signal.
Finally, the multilingual PLM is continuously pretrained on synthetic dataset with multi-task learning manner.
Besides, our proposed models, named \cmodel-MRA and \cmodel-UP respectively, can benefit from finetuning with human-annotated data over three tasks at once, not requiring extra task-specific training.

Experimental results demonstrate the superiority of \cmodel.
Compared to various strong baseline systems on each task, \cmodel, which unifies \cref, \csrc~and \csrcref~tasks into one \textit{single model}, achieves consistently absolute improvements of Kendall's $\tau$ correlations at 1.1, 2.3 and 1.1 scores on English-targeted translation directions of WMT 2019 Metric Shared task~\cite{fonseca2019findings}, respectively.
Meanwhile, after introducing multilingual-targeted support for our unified pretraining strategy, a single model named \cmodel-MUP also gives dominant results against  existing methods on  non-English-targeted translation evaluation tasks. 
Furthermore, our method can also achieve competitive results over WMT 2020 QE task compared with the winner submission~\cite{ranasinghe2020transquest}.
Ablation studies reveal that, the proposed MRA and unified pretraining strategies are both important for model performance, making the model preserve the outstanding performance and multi-task transferability concurrently.

\section{Related Work}
In this section, we briefly introduce the three directions of translation evaluation.

\subsection{Reference-Only Evaluation}
\cref~assesses the translation quality via comparing the translation candidate and the given reference. 
In this setting, the two inputs are written in the same language, thus being easily applied in most of the metric tasks. 
In the early stages, statistical methods are dominant solutions due to their strengths in wide language support and intuitive design. 
These methods measure the surface text similarity for a range of linguistic features, including  n-gram~\citep[BLEU,][]{papineni2002bleu}, token~\citep[TER,][]{snover2006a}, and character~\citep[ChrF \& ChrF++,][]{popovic2015chrf,popovic2017chrf++}.
However, recent studies pointed out that these metrics have low consistency with human judgments and insufficiently evaluate high-qualified MT systems~\cite{freitag2020bleu,rei2020comet,mathur2020tangled}.

Consequently, with the rapid development of PLMs, researchers have been paying their attention to model-based approaches.
The basic idea of these studies is to collect sentence representations for similarity calculation~\citep[BERTScore,][]{zhang2020bertscore} or evaluating probabilistic confidence~\citetext{\citealp[PRISM-ref,][]{thompson2020automatic}; \citealp[BARTScore,][]{yuan2021bartscore}}. 
To further improve the model, \newcite{sellam2020bleurt} pretrained a specific PLM for the translation evaluation (BLEURT), while \newcite{lo2019yisi} combined statistical and representative features (YiSi-1). Both these methods achieve higher correlations with human judgments than statistical counterparts.



\subsection{Source-Only Evaluation}
\csrc , which also refers to quality estimation\footnote{Refer to ``quality estimation'' or ``reference-free metric'' in  WMT  (\href{http://www.statmt.org/wmt19/qe-task.html}{http://www.statmt.org/wmt19/qe-task.html}, \href{http://www.statmt.org/wmt21/metrics-task.html}{http://www.statmt.org/wmt21/metrics-task.html}).},  is an important  translation evaluation task especially for the scenario where the ground-truth reference is unavailable.
It takes the source-side sentence and the translation candidate as inputs for the quality estimation.
To achieve this, the methods are required to model cross-lingual semantic alignments. 
Similar to reference-only evaluation, statistical-based~\citep{ranasinghe2020transquest},  model-based~\citetext{\citealp[TransQuest,][]{ranasinghe2020transquest};~\citealp[PRISM-src,][]{thompson2020automatic}}, and feature combination~\citetext{\citealp[YiSi-2,][]{lo2019yisi}} are typical and advanced methods in this tasks.

\subsection{Source-Reference-Combined Evaluation}
Aside from the above tasks that only consider either source or target side at one time, \csrcref~takes both  source and reference sentences into account. 
In this way, methods in this context can evaluate the translation candidate via utilizing the features from both sides. 
As a rising paradigm among translation  evaluation tasks, \csrcref~also profits from the development of cross-lingual PLMs. For example, 
finetuning PLMs over human-annotated datasets~\citep[COMET,][]{rei2020comet} achieves new state-of-the-art results among all evaluation approaches in WMT 2020~\cite{mathur2020results}.

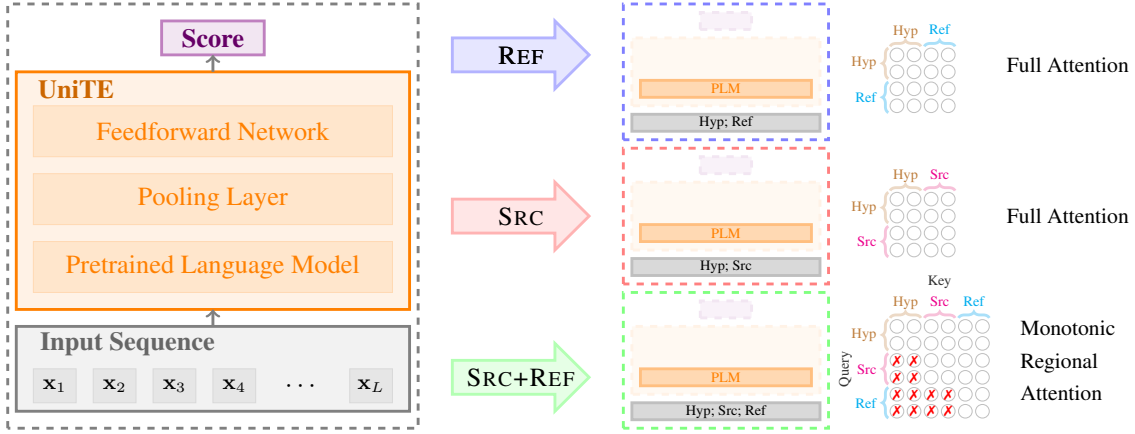
\begin{figure*}
    \centering
    \scalebox{0.9}
    {
        \begin{tikzpicture}
            
            \draw[rectangle, draw=orange, fill=orange!10, very thick] (-2.875, 0) rectangle node{} (2.875, 3.5);
            \draw[rectangle, color=orange, draw=orange!30, fill=orange!20] (-2.625, 0.25) rectangle node{Pretrained Language Model} (2.625, 1.0);
            \draw[rectangle, color=orange, draw=orange!30, fill=orange!20] (-2.625, 1.25) rectangle node{Pooling Layer} (2.625, 2.0);
            \draw[rectangle, color=orange, draw=orange!30, fill=orange!20] (-2.625, 2.25) rectangle node{Feedforward Network} (2.625, 3.0);
            
            \draw[color=orange!80!black] (-2, 3.25) node{\textbf{\cmodel}};
            
            \draw[rectangle, draw=gray, fill=gray!10, very thick] (-2.875, -1.5) rectangle node(){} (2.875, -0.25);
            
            \draw[rectangle, draw=gray!30, fill=gray!20] (-2.625, -1.375) rectangle node(){\small{$\mathbf{x}_1$}} (-2.0, -0.875);
            
            \draw[rectangle, draw=gray!30, fill=gray!20] (-1.75, -1.375) rectangle node(){\small{$\mathbf{x}_2$}} (-1.125, -0.875);
            
            \draw[rectangle, draw=gray!30, fill=gray!20] (-0.875, -1.375) rectangle node(){\small{$\mathbf{x}_3$}} (-0.25, -0.875);
            
            \draw[rectangle, draw=gray!30, fill=gray!20] (0.0, -1.375) rectangle node(){\small{$\mathbf{x}_4$}} (0.625, -0.875);
            
            \draw (1.3125, -1.125) node{$\cdots$};
            
            \draw[rectangle, draw=gray!30, fill=gray!20] (2.0, -1.375) rectangle node(){\small{$\mathbf{x}_L$}} (2.625, -0.875);
            
            \draw[color=gray!80!black] (-1.25, -0.5) node{\textbf{Input Sequence}};
            
            \draw[rectangle, color=violet!80!black, draw=violet!50, fill=violet!10, very thick] (-0.75, 3.75) rectangle node{\textbf{Score}} (0.75, 4.25);
            
            \draw [->, color=black!50, very thick](0, -0.25) -- (0, 0.0);
            
            \draw [->, color=black!50, very thick](0, 3.5) -- (0, 3.75);
            
            \draw [rectangle, draw=black!50, dashed, very thick] (-3.0, -1.75) rectangle node{} (3.0, 4.5);
            
            \filldraw[draw=blue!30, fill=blue!10, very thick] (3.5, 4.0) -- (5.0, 4.0) -- (5.0, 4.25) -- (5.5, 3.75) -- (5.0, 3.25) -- (5.0, 3.5) -- (3.5, 3.5) -- cycle;
            \draw (4.5, 3.75) node{\textsc{Ref}};
            
            \filldraw[draw=red!30, fill=red!10, very thick] (3.5, 1.625) -- (5.0, 1.625) -- (5.0, 1.875) -- (5.5, 1.375) -- (5.0, 0.875) -- (5.0, 1.125) -- (3.5, 1.125) -- cycle;
            \draw (4.5, 1.375) node{\textsc{Src}};
            
            \filldraw[draw=green!30, fill=green!10, very thick] (3.5, -0.75) -- (5.0, -0.75) -- (5.0, -0.5) -- (5.5, -1.0) -- (5.0, -1.5) -- (5.0, -1.25) -- (3.5, -1.25) -- cycle;
            \draw (4.5, -1) node{\textsc{Src+Ref}};

            \filldraw[draw=blue!30, fill=blue!10, very thick] ;
            \draw (4.5, 3.75) node{\textsc{Ref}};
            
            \draw[rectangle, draw=blue!50, very thick, dashed] (6.0, 2.5) rectangle node(){} (9.0, 4.5);
            
            \draw[rectangle, draw=orange!10, fill=orange!5, very thick, dashed] (6.125, 3.0) rectangle node{} (8.875, 4.0);
            
            \draw[rectangle, draw=gray!50, fill=gray!30, very thick] (6.125, 2.625) rectangle node{\tiny{Hyp; Ref}}(8.875, 2.875);
            
            \draw[rectangle, draw=violet!10, fill=violet!5, very thick, dashed] (7.125, 4.125) rectangle node{} (7.875, 4.375);
            
            \draw[rectangle, color=orange, draw=orange!50, fill=orange!30, very thick] (6.25, 3.125) rectangle node{\tiny{PLM}} (8.75, 3.375);
            
            \foreach \x in {0,...,3}
                \foreach \y in {0,...,3}
                {
                    \draw[draw=gray!50, fill=white] (10+0.25*\x, 3+0.25*\y) circle (0.1);
                }
            
            \draw [color=brown, draw=brown!30, decorate, decoration = {brace}, very thick] (9.9, 3.875) -- (10.35, 3.875) node[pos=0.5, above=-0.025]{\tiny{Hyp}};
            \draw [color=cyan, draw=cyan!30, decorate, decoration = {brace}, very thick] (10.4, 3.875) -- (10.85, 3.875) node[pos=0.5, above=0.025]{\tiny{Ref}};
            
            \draw [color=brown, draw=brown!30, decorate, decoration = {brace}, very thick] (9.875, 3.4) -- (9.875, 3.85) node[pos=0.5, left=0.025]{\tiny{Hyp}};
            \draw [color=cyan, draw=cyan!30, decorate, decoration = {brace}, very thick] (9.875, 2.9) -- (9.875, 3.35) node[pos=0.5, left=0.025]{\tiny{Ref}};
    
            \draw [color=black] (12.5, 3.6) node{\small{Full Attention}};
    
            \draw[rectangle, draw=red!50, very thick, dashed] (6.0, 0.375) rectangle node(){} (9.0, 2.375);
            
            \draw[rectangle, draw=orange!10, fill=orange!5, very thick, dashed] (6.125, 0.875) rectangle node{} (8.875, 1.875);
            
            \draw[rectangle, draw=gray!50, fill=gray!30, very thick] (6.125, 0.5) rectangle node{\tiny{Hyp; Src}}(8.875, 0.75);
            
            \draw[rectangle, draw=violet!10, fill=violet!5, very thick, dashed] (7.125, 2.0) rectangle node{} (7.875, 2.25);
            
            \draw [color=black!80] (10.625, 0.4) node{\tiny{Key}};
            \draw [color=black!80] (9.25, -0.875) node{\rotatebox{90}{\tiny{Query}}};

            \draw[rectangle, color=orange, draw=orange!50, fill=orange!30, very thick] (6.25, 1.0) rectangle node{\tiny{PLM}} (8.75, 1.25);
            
            \foreach \x in {0,...,3}
                \foreach \y in {0,...,3}
                {
                    \draw[draw=gray!50, fill=white] (10+0.25*\x, 0.875+0.25*\y) circle (0.1);
                }
            
            \draw [color=brown, draw=brown!30, decorate, decoration = {brace}, very thick] (9.9, 1.75) -- (10.35, 1.75) node[pos=0.5, above=-0.025]{\tiny{Hyp}};
            \draw [color=magenta, draw=magenta!30, decorate, decoration = {brace}, very thick] (10.4, 1.75) -- (10.85, 1.75) node[pos=0.5, above=0.025]{\tiny{Src}};
            
            \draw [color=brown, draw=brown!30, decorate, decoration = {brace}, very thick] (9.875, 1.275) -- (9.875, 1.725) node[pos=0.5, left=0.025]{\tiny{Hyp}};
            \draw [color=magenta, draw=magenta!30, decorate, decoration = {brace}, very thick] (9.875, 0.775) -- (9.875, 1.225) node[pos=0.5, left=0.025]{\tiny{Src}};
            
            \draw [color=black] (12.5, 1.375) node{\small{Full Attention}};
    
            \draw[rectangle, draw=green!50, very thick, dashed] (6.0, -1.75) rectangle node(){} (9.0, 0.25);
            
            \draw[rectangle, draw=orange!10, fill=orange!5, very thick, dashed] (6.125, -1.25) rectangle node{} (8.875, -0.25);
            
            \draw[rectangle, draw=gray!50, fill=gray!30, very thick] (6.125, -1.625) rectangle node{\tiny{Hyp; Src; Ref}}(8.875, -1.375);
            
            \draw[rectangle, draw=violet!10, fill=violet!5, very thick, dashed] (7.125, -0.125) rectangle node{} (7.875, 0.125);
            
            \draw[rectangle, color=orange, draw=orange!50, fill=orange!30, very thick] (6.25, -1.125) rectangle node{\tiny{PLM}} (8.75, -0.875);
            
            \foreach \x in {0,...,5}
                \foreach \y in {0,...,5}
                {
                    \ifthenelse{\(\x<2 \AND \y>1 \AND \y<4\) \OR \(\x<4 \AND \y<2\)}
                    {
                        \draw[draw=gray!50, fill=white] (10+0.25*\x, -1.5+0.25*\y) circle (0.1) node {\textcolor{red}{\tiny{\ding{55}}}};
                    }
                    {
                        \draw[draw=gray!50, fill=white] (10+0.25*\x, -1.5+0.25*\y) circle (0.1);
                    }
                }
            
            \draw [color=brown, draw=brown!30, decorate, decoration = {brace}, very thick] (9.9, -0.125) -- (10.35, -0.125) node[pos=0.5, above=-0.025]{\tiny{Hyp}};
            \draw [color=magenta, draw=magenta!30, decorate, decoration = {brace}, very thick] (10.4, -0.125) -- (10.85, -0.125) node[pos=0.5, above=0.025]{\tiny{Src}};
            \draw [color=cyan, draw=cyan!30, decorate, decoration = {brace}, very thick] (10.9, -0.125) -- (11.35, -0.125) node[pos=0.5, above=0.025]{\tiny{Ref}};
            
            \draw [color=brown, draw=brown!30, decorate, decoration = {brace}, very thick] (9.875, -0.6) -- (9.875, -0.15)  node[pos=0.5, left=0.025]{\tiny{Hyp}};
            \draw [color=magenta, draw=magenta!30, decorate, decoration = {brace}, very thick] (9.875, -1.1) -- (9.875, -0.65) node[pos=0.5, left=0.025]{\tiny{Src}};
            \draw [color=cyan, draw=cyan!30, decorate, decoration = {brace}, very thick] (9.875, -1.6) -- (9.875, -1.15) node[pos=0.5, left=0.025]{\tiny{Ref}};
            
            \node[draw, color=black, draw=none, fill=none, align=left]  at (12.5, -0.75) () {\small{Monotonic} \\ \small{Regional} \\ \small{Attention}};

        \end{tikzpicture}
    }
    \caption{Illustration of \cmodel. Our model can give predictions for different data items formatted as \fboxsep0.5ex\cref, \fboxsep0.5ex\csrc, or \fboxsep0.5ex\csrcref~setting, unifying all evaluation tasks into one single model without additional modifications. For \csrcref, we show the hard design for monotonic regional attention. \textcolor{red}{\ding{55}} denotes the masked attention logits.}
    \label{fig.architecture}
\end{figure*}

\section{Methodology}
As mentioned above, massive methods are proposed for different automatic evaluation tasks.
On the one hand, it is inconvenient and expensive to develop and employ different metrics for different evaluation scenarios.
On the other hand, separate models absolutely overlook the commonalities among these evaluation tasks, of which knowledge potentially benefits all three tasks.
In order to fulfill the aim of unifying the functionalities on \cref, \csrc, and \csrcref~into one model, in this section, we introduce \textbf{\cmodel} (Figure~\ref{fig.architecture}).

\subsection{Model Architecture}
\label{sec:model}

By receiving a data example composing of hypothesis, source, and reference segment, \cmodel~first modifies it into concatenated sequence following the given setting as \cref, \csrc, or \csrcref:
\begin{align}
    \mathbf{x}_\textsc{Ref} & = \texttt{Concat} (\mathbf{h}, \mathbf{r}) \in \mathbb{R}^{(l_h + l_r)}, \notag \\
    \mathbf{x}_\textsc{Src} & = \texttt{Concat}(\mathbf{h}, \mathbf{s}) \in \mathbb{R}^{(l_h + l_s)},  \\
    \mathbf{x}_\textsc{Src+Ref} & = \texttt{Concat}(\mathbf{h}, \mathbf{s}, \mathbf{r}) \in \mathbb{R}^{(l_h + l_s + l_r)}, \notag
\end{align}
where $\mathbf{h}$, $\mathbf{s}$ and $\mathbf{r}$ are hypothesis, source and reference segments, with the corresponding sequence lengths being $l_h$, $l_s$ and $l_r$, respectively.
The input sequence is then fed to PLM to derive representations $\tilde{\mathbf{H}}$.
Take \cref~as an example:
\begin{align}
    \tilde{\mathbf{H}}_\textsc{Ref} = \texttt{PLM}(\mathbf{x}_\textsc{Ref}) \in \mathbb{R}^{(l_h + l_r) \times d},
\end{align}
where $d$ is the model size of PLM.
According to~\newcite{ranasinghe2020transquest}, we use the first output representation as the input of feedforward layer. 

Compared to existing methods~\cite{zhang2020bertscore,rei2020comet} which take sentence-level representations for evaluation, the advantages of our architecture design are as follows.
First, our \cmodel~model can benefit from layer-coordinated semantical interactions inside every one of PLM layers, which is proven effective on capturing diverse linguistic features ~\cite{he2018layer,lin2019open,jawahar2019what,tenney2019bert,rogers2020primer}.
Second, for the unified approach of our model, the concatenation provides the unifying format for all task inputs, turning our model into a more general architecture.
When conducting different evaluation tasks, our model requires no further modification inside.
Note here, to keep the consistency across all evaluation tasks, as well as ease the unified learning, $\mathbf{h}$ is always located at the beginning of the input sequence.

After deriving $\tilde{\mathbf{H}}_{\textsc{Ref}}$, a pooling block is arranged after PLM which gives sequence-level representations $\mathbf{H}_{\textsc{Ref}}$.
Finally, a feedforward network takes $\mathbf{H}_{\textsc{Ref}}$ as input, and gives a scalar $p$ as prediction:
\begin{align}
    \mathbf{H}_\textsc{Ref} & = \texttt{Pool}(\tilde{\mathbf{H}}_\textsc{Ref}) \in \mathbb{R}^{d}, \\
    p_\textsc{Ref} & = \texttt{FeedForward}(\mathbf{H}_\textsc{Ref}) \in \mathbb{R}^{1}.
\end{align}
For training, we encourage the model to reduce the mean squared error with respect to given score $q$:
\begin{align}
    \mathcal{L}_\textsc{Ref} = (p_\textsc{Ref} - q) ^ 2.
\end{align}

However, for the pretraining of most PLMs~\citep[e,g., XLM-R,][]{conneau2020unsupervised}, the input patterns are designed to receive two segments at most.
Thus there exists a gap between the pretraining of PLM and the joint training of \cmodel~where the concatenation of three fragments is used as input.
Moreover, previous study~\cite{takahashi2020automatic} shows that directly training over \csrcref~by following such design leads to worse performance than \cref~scenario.
To alleviate this issue, we propose two strategies: \textbf{Monotonic Regional Attention} as described in~\S\ref{sec:mra} and \textbf{Unified Pretraining} in~\S\ref{sec:up}.

\subsection{Monotonic Regional Attention}
\label{sec:mra}
To fill the modeling gap between the pretraining of PLM and the joint training of three downstream tasks, a natural idea is to unify the number of involved segments when modeling semantics for \csrc, \cref~and \csrcref~tasks.
Following this, we propose to modify the attention mask of \csrcref~to simulate the modeling of two segments in \csrc~and \cref.
Specifically, when calculating the attention logits, semantics from a specific segment are only allowed to derive information from two segments at most.
Considering the conventional attention module:
\begin{align}
    \mathbf{A} = \mathrm{Softmax}(\frac{\mathbf{Q}\mathbf{K}^\top}{\sqrt{d}}) \in \mathbb{R}^{L \times L},
\end{align}
where $L$ is the sequential length for input, $\mathbf{Q}, \mathbf{K} \in \mathbb{R}^{L \times d}$ are query and key representations, respectively.\footnote{For simplicity, we omit the multi-head mechanism.}
As to monotonic regional attention (MRA), we simply add a mask $\mathbf{M}$ to the softmax logits to control attention flows:
\begin{align}
    \mathbf{A} & = \mathrm{Softmax}(\frac{\mathbf{Q}\mathbf{K}^\top}{\sqrt{d}} + \mathbf{M}) \in \mathbb{R}^{L \times L}, \\
    \mathbf{M}_{ij} & = 
        \begin{cases}
            -\infty & (i,j) \in \mathbf{U}, \\
            0 & \text{otherwise},
        \end{cases}
\end{align}
where $\mathbf{U}$ stores the index pairs of all masked areas. 

Following this idea, the key of MRA is how to design the matrix $\mathbf{U}$.
For the cases where interactions inside each segment, we believe that these self-interactions are beneficial to the modeling.
For other cases where interactions are arranged across segments, three patterns are included: hypothesis-reference, source-reference, and hypothesis-source.
Intuitively, the former two parts are beneficial for model training, since they might contribute the monolingual signals and cross-lingual disambiguation to evaluation, respectively.
This leaves the only case, where our experimental analysis also verifies (see \S\ref{sec.ablation.mra}), that interaction between hypothesis and source leads to the performance decrease for \csrcref~task, thus troubling the unifying.

\tikzset{
semi/.style={
  semicircle,
  draw,
  }
}

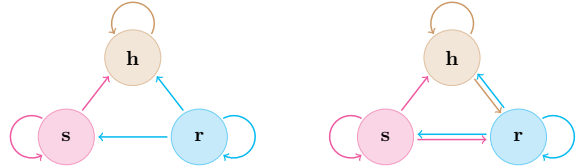
\begin{figure}
    
    \centering
    \scalebox{0.7}{
        \begin{tikzpicture}
        {

            \node[fill=brown!20, draw=brown!50, circle, minimum size=30pt] (hyp) at (0, 0) {$\mathbf{h}$};
            \node[fill=magenta!20, draw=magenta!50, circle, minimum size=30pt] (src) at (-1.25, -1.5) {$\mathbf{s}$};
            \node[fill=cyan!20, draw=cyan!50, circle, minimum size=30pt] (ref) at (1.25, -1.5) {$\mathbf{r}$};
            
            \draw[brown!80, thick, ->] (0.35, 0.45) arc (-30:210:0.4);
            \draw[magenta!80, thick, ->] (-1.7, -1.15) arc (60:300:0.4);
            \draw[cyan!80, thick, ->] (1.7, -1.15) arc (120:-120:0.4);
            
            \draw[cyan!80, thick, ->] (0.65, -1.5) -- (-0.65, -1.5);
            \draw[cyan!80, thick, ->] (0.95, -1.0) -- (0.45, -0.35);
            \draw[magenta!80, thick, ->] (-0.95, -1.0) -- (-0.45, -0.35);

            \node[fill=brown!20, draw=brown!50, circle, minimum size=30pt] (hyp) at (6, 0) {$\mathbf{h}$};
            \node[fill=magenta!20, draw=magenta!50, circle, minimum size=30pt] (src) at (4.75, -1.5) {$\mathbf{s}$};
            \node[fill=cyan!20, draw=cyan!50, circle, minimum size=30pt] (ref) at (7.25, -1.5) {$\mathbf{r}$};
            
            \draw[brown!80, thick, ->] (6.35, 0.45) arc (-30:210:0.4);
            \draw[magenta!80, thick, ->] (4.3, -1.15) arc (60:300:0.4);
            \draw[cyan!80, thick, ->] (7.7, -1.15) arc (120:-120:0.4);
            
            \draw[cyan!80, thick, ->] (6.65, -1.45) -- (5.35, -1.45);
            \draw[magenta!80, thick, ->] (5.35, -1.55) -- (6.65, -1.55);
            
            \draw[cyan!80, thick, ->] (6.98, -0.95) -- (6.48, -0.3);
            \draw[brown!80, thick, ->] (6.42, -0.4) -- (6.92, -1.05);
            
            \draw[magenta!80, thick, ->] (5.05, -1.0) -- (5.55, -0.35);
        }
            
        \end{tikzpicture}
    }
    
    \caption{Attention flows in monotonic regional attention. $\mathbf{h}$, $\mathbf{s}$ and $\mathbf{r}$ are hypothesis, source and reference, respectively. We prevent specified interactions in \fboxsep0.5ex\colorbox{green!10}{\textsc{Src+Ref}}~training via modifying the attention mask with regional properties. We show the hard (left) and soft design (right, no $\mathbf{h}\rightarrow\mathbf{s}$) in this figure.}
    \label{fig.monotonic}
\end{figure}
To give more fine-grained designs, we propose two approaches for \cmodel-MRA, which apply the MRA mechanism into \cmodel~model (Figure~\ref{fig.monotonic}):

\begin{itemize}
    \item Hard MRA. Only monotonic attention flows are allowed. Interactions between any two segments are strictly unidirectional through the entire PLM, where
    $\mathbf{U}$ stores the index pairs of unidirectional interactions of $\mathbf{h}\rightarrow\mathbf{r}$, $\mathbf{s}\rightarrow\mathbf{r}$ and $\mathbf{h}\rightarrow\mathbf{s}$, where ``$\rightarrow$'' denotes the direction of attention flows.
    \item Soft MRA. Specific attention flows are forbidden inside each attention module.
    The involved two segments may interact inside a higher layer.
    In practice, index pairs which denoting $\mathbf{h}\rightarrow\mathbf{s}$ or $\mathbf{s}\rightarrow\mathbf{h}$ between source and hypothesis are stored in $\mathbf{U}$.
\end{itemize}

Note that, although the processing in source and reference may be affected because their positions are not indexed from the start, related studies on positional embeddings reveal that, PLM can well capture relative positional information~\cite{wang2020what}, which dispels this concern.

\subsection{Unified Pretraining}
\label{sec:up}
To further bridge the modeling gap between PLM and the joint training of \cmodel~mentioned in~\S\ref{sec:model}, we propose a unified pretraining strategy including the following main stages: 1) collecting and downgrading synthetic data; 2) labeling examples with a novel ranking-based strategy; 3) multi-task learning for unified pretraining and finetuning.

\paragraph{Synthetic Data Collection}
As our approach aims at evaluating the quality of translations, generated hypotheses with NMT models are ideal synthetic data.
To further improve the diversity of synthetic data quality,
we follow existing experiences~\cite{sellam2020bleurt,wan2021robleurt} to apply the word and span dropping strategy to downgrade a portion of hypotheses.
The collected data totally contains $N$ triplets composing of hypothesis, source and reference segments, which is formed as $\mathcal{D}' =  \{\langle\mathbf{h}^i, \mathbf{s}^i, \mathbf{r}^i\rangle\}_{i=1}^{N}$.

\paragraph{Data Labeling}
After obtaining the synthetic data, the next step is to augment each data pair with a label which serves as the signal of unified pretraining.
To stabilize the model training, as well as normalize the distributions across all score systems and languages, we propose a novel ranking-based approach.
This method is based on the idea of Borda count~\cite{ho1994decision,emerson2013original}, which provides more precise and well-distributed synthetic data labels than Z-score normalization.

Specifically, we first use available approaches to derive the predicted score $\hat{q}^i$ for each item, yielding labeled synthetic quadruple examples formed as $\mathcal{D}'' =  \{\langle\mathbf{h}^{i}, \mathbf{s}^{i}, \mathbf{r}^{i}, \hat{q}^i\rangle\}_{i=1}^{N}$.
Then, we tag each example with its rank index $\tilde{q}^i$ referring to $\hat{q}^i$:
\begin{align}
    \tilde{q}^i = \texttt{IndexOf}(\hat{q}^i, \mathcal{Q}),
\end{align}
where $\mathcal{Q}$ is the list storing all the sorted $\hat{q}^i$ descendingly.
Then, we use the conventional Z-score strategy to normalize the scores:
\begin{align}
    q^i = \frac{\tilde{q}^i - \mu}\sigma,
\end{align}
where $\mu$ and $\sigma$ are the mean and the standard deviation of values in $\mathcal{Q}$, respectively.
The dataset thus updates its format to $\mathcal{D} =  \{\langle\mathbf{h}^i, \mathbf{s}^i, \mathbf{r}^i, q^i\rangle\}_{i=1}^{N}$.
Note here that, an example with higher $\hat{q}^i$ is assigned with higher $\tilde{q}^i$, thus a larger value of $q^i$.

Compared to related approaches which apply Z-score normalization~\cite{bojar2018findings}, or leave the conventional labeled scores as signals for learning~\citep[\textit{i.e.}, knowledge distillation,][]{kim2016sequence,phuong2019towards}, our approach can alleviate the bias of chosen model for labeling and prior distributional disagreement of scores.
For example, different methods may give scores with different distributions.
Especially for translation directions of low-resource, scores may follow skewed distribution~\cite{sellam2020bleurt}, which has a disagreement with rich-resource scenarios.
Our method can unify the distribution of all labeling data into the same scale, which can also be easily applied by the ensembling strategy.

\paragraph{Multi-task Pretrainig and Finetuning}
To unify all evaluation scenarios into one model, we apply multi-task learning for both pretraining and finetuning.
For each step, we arrange three substeps for all input formats, yielding $\mathcal{L}_\textsc{Ref}$, $\mathcal{L}_\textsc{Src}$, and $\mathcal{L}_\textsc{Src+Ref}$, respectively.
The final learning objective is to reduce the summation of all losses:
\begin{align}
    \mathcal{L} = \mathcal{L}_\textsc{Ref} + \mathcal{L}_\textsc{Src} + \mathcal{L}_\textsc{Src+Ref}.
\end{align}

\begin{table*}[t]
    \small
    \centering
    \scalebox{1.0}
    {
        \begin{tabular}{lcccccccc}
            \toprule
            
            \multirow{2}{*}{\textbf{Model}} & \multicolumn{4}{c}{\textbf{High-resource}} & \multicolumn{3}{c}{\textbf{Zero-shot}} & \multirow{2}{*}{\textbf{Avg.}} \\
            \cmidrule(l{2pt}r{2pt}){2-5}\cmidrule(l{2pt}r{2pt}){6-8}
            & \textit{\underline{De-En}} & \textit{\underline{Ru-En}} & \textit{\underline{Zh-En}} & \underline{Fi-En} & Gu-En & Kk-En & Lt-En & \\
            
            
            \midrule
            
            \multicolumn{9}{c}{\textit{Reference-only Evaluation}} \\
            \cdashline{1-9}\noalign{\vskip 0.1ex}
            
            $^\heartsuit$BLEU~\cite{papineni2002bleu} & ~~5.4 & 11.5 & 32.1 & 23.6 & 19.4 & 27.6 & 24.9 & \colorbox{blue!10}{20.6} \\
            $^\spadesuit$ChrF~\cite{popovic2015chrf} & 12.3 & 17.7 & 37.1 & 29.2 & 24.0 & 32.3 & 30.4 & \colorbox{blue!10}{26.1} \\
            $^\heartsuit$BERTScore~\cite{zhang2020bertscore} & 19.0 & 22.1 & 43.0 & 35.4 & 29.2 & 35.1 & 38.1 & \colorbox{blue!10}{31.7} \\
            $^\heartsuit$BLEURT~\cite{sellam2020bleurt} & 17.4 & 22.0 & 43.6 & 37.4 & 31.3 & 37.2 & 38.8 & \colorbox{blue!10}{32.5} \\
            $^\spadesuit$YiSi-1~\cite{lo2019yisi} & 16.4 & 21.7 & 42.6 & 34.7 & 31.2 & \textbf{44.0} & 37.6 & \colorbox{blue!10}{32.6} \\
            $^\heartsuit$PRISM-ref~\cite{thompson2020automatic} & 20.4 & \textbf{22.5} & 43.8 & 35.7 & 31.3 & 43.4 & 38.2 & \colorbox{blue!10}{33.6} \\
            $^\heartsuit$BARTScore~\cite{yuan2021bartscore} & 23.8 & 21.9 & 44.7 & 37.4 & 31.8 & 37.6 & 38.6 & \colorbox{blue!10}{33.7} \\
            $^\diamondsuit$XLM-R+Concat~\cite{takahashi2020automatic} & 24.5 & 21.8 & 45.8 & 37.0 & 31.5 & 37.4 & 39.5 & \colorbox{blue!10}{33.9} \\
            $^\diamondsuit$RoBERTa+Concat~\cite{takahashi2020automatic} & 25.1 & 22.4 & 46.4 & 36.2 & 30.8 & 38.0 & \textbf{40.0} & \colorbox{blue!10}{34.1} \\
            \cdashline{1-9}\noalign{\vskip 0.1ex}
            
            \cmodel-MRA & 25.2 & 22.4 & 46.4 & 36.5 & 31.6 & 38.4 & 39.1 & \colorbox{blue!10}{34.2} \\
            \cmodel-UP & \textbf{25.9} & 21.9 & \textbf{46.7} & \textbf{37.9} & \textbf{32.2} & 38.7 & \textbf{40.0} & \colorbox{blue!10}{\textbf{34.8}} \\
            
            \midrule
            
            \multicolumn{9}{c}{\textit{Source-only Evaluation}} \\
            \cdashline{1-9}\noalign{\vskip 0.1ex}
            
            $^\spadesuit$YiSi-2~\cite{lo2019yisi} & ~~6.8 & ~~5.3 & 25.3 & 12.6 & ~-0.1 & ~~9.6 & ~~7.5 & \colorbox{red!10}{~~9.5} \\
            $^\heartsuit$PRISM-src~\cite{thompson2020automatic} & 10.9 & \textbf{17.8} & 33.6 & 30.0 & 10.2 & \textbf{39.1} & 35.6 & \colorbox{red!10}{25.3} \\
            $^\heartsuit$MTransQuest~\cite{ranasinghe2020transquest} & 11.1 & 14.0 & 32.1 & 29.7 & 27.2 & 31.6 & 30.7 & \colorbox{red!10}{25.2} \\
            $^\diamondsuit$MTransQuest~\cite{ranasinghe2020transquest} & 17.0 & 17.3 & 37.6 & 29.2 & 26.5 & 31.9 & 34.2 & \colorbox{red!10}{27.7} \\
            $^\diamondsuit$XLM-R+Concat~\cite{takahashi2020automatic} & 16.9 & 17.6 & 38.1 & 29.1 & 26.2 & 31.6 & 34.3 & \colorbox{red!10}{27.7} \\
            \cdashline{1-9}\noalign{\vskip 0.1ex}
            
            \cmodel-MRA & 17.4 & 17.7 & 41.0 & \textbf{34.3} & 29.0 & 32.7 & \textbf{36.2} & \colorbox{red!10}{29.7} \\
            \cmodel-UP & \textbf{19.3} & 16.9 & \textbf{41.4} & 34.0 & \textbf{29.7} & 33.6 & 35.4 & \colorbox{red!10}{\textbf{30.0}} \\
            
            \midrule
            
            \multicolumn{9}{c}{\textit{Source-Reference-Combined Evaluation}} \\
            \cdashline{1-9}\noalign{\vskip 0.1ex}
            
            $^\diamondsuit$XLM-R+Concat~\cite{takahashi2020automatic} & 24.0 & 22.0 & 44.7 & 35.7 & 30.4 & 37.2 & 38.9 & \colorbox{green!10}{33.4} \\
            $^\diamondsuit$COMET~\cite{rei2020comet} & 23.4 & 20.7 & 45.8 & 36.2 & 30.9 & 37.9 & 40.3 & \colorbox{green!10}{33.6} \\
            \cdashline{1-9}\noalign{\vskip 0.1ex}
            
            \cmodel-MRA & 25.6 & \textbf{22.9} & 46.9 & 37.6 & 31.6 & 38.5 & \textbf{40.5} & \colorbox{green!10}{34.8} \\
            \cmodel-UP & \textbf{26.0} & 22.0 & \textbf{47.2} & \textbf{37.7} & \textbf{32.3} & \textbf{39.4} & 40.0 & \colorbox{green!10}{\textbf{35.0}} \\

            \bottomrule
        \end{tabular}
    }
    \caption{Kendall's Tau correlation (\%) results on English-targeted language pairs of WMT 2019 Metrics Task test set. \textit{Italic} and \underline{underlined} translation directions indicate that corresponding data items are available in pretraining and finetuning training set, respectively. Baselines marked with $^\heartsuit$, $^\spadesuit$ and $^\diamondsuit$ mean that scores are derived from official release, WMT official report~\cite{ma2019results}, and our reimplementation, respectively. Colored background indicates that evaluation follows \fboxsep0.5ex\cref, \fboxsep0.5ex\csrc~and \fboxsep0.5ex\csrcref~setting. Best viewed in \textbf{bold}.}
    \label{table.main_results_metric_x-en}
\end{table*}

\begin{table*}[t]
    \centering
    \scalebox{0.675}{
        \begin{tabular}{lcccccccccccc}
            \toprule
            
            
            \multirow{2}{*}{\textbf{Model}} & \multicolumn{5}{c}{\textbf{High-resource}} & \multicolumn{6}{c}{\textbf{Zero-shot}} & \multirow{2}{*}{\textbf{Avg.}} \\
            \cmidrule(l{2pt}r{2pt}){2-6}\cmidrule(l{2pt}r{2pt}){7-12}
             & \textit{\underline{En-Cs}} & \textit{\underline{En-De}} & \textit{\underline{En-Ru}} & \textit{\underline{En-Zh}} & \underline{En-Fi} & En-Gu & En-Kk & En-Lt & De-Cs & De-Fr & Fr-De & \\
            
            \midrule
            
            \multicolumn{13}{c}{\textit{Reference-only Evaluation}} \\
            \cdashline{1-13}\noalign{\vskip 0.1ex}
            
            
            $^\heartsuit$BLEU~\cite{papineni2002bleu} & 36.4 & 24.8 & 46.9 & 23.5 & 39.5 & 46.3 & 36.3 & 33.3 & 22.2 & 22.6 & 17.3 & \colorbox{blue!10}{31.7} \\
            $^\spadesuit$ChrF~\cite{popovic2015chrf} & 44.4 & 32.1 & 54.8 & 24.1 & 51.8 & 54.8 & 51.0 & 43.8 & 34.1 & 28.7 & 27.4 & \colorbox{blue!10}{40.6} \\
            $^\heartsuit$BERTScore~\cite{zhang2020bertscore} & 50.0 & 36.3 & \textbf{58.5} & 35.6 & 52.7 & 56.8 & 54.0 & 46.4 & 35.8 & 32.9 & 30.0 & \colorbox{blue!10}{44.5} \\
            $^\spadesuit$YiSi-1~\cite{lo2019yisi} & 47.5 & 35.1 & \textbf{58.5} & 35.5 & 53.7 & 55.1 & 54.6 & 47.0 & 37.6 & 34.9 & 31.0 & \colorbox{blue!10}{44.6} \\
            $^\spadesuit$BLEURT~\cite{sellam2020learning} & 60.3 & 42.2 & 49.2 & 33.7 & 61.5 & 57.7 & 55.8 & 58.4 & 46.1 & 44.9 & \textbf{42.7} & \colorbox{blue!10}{50.2} \\
            $^\diamondsuit$XLM-R+Concat~\cite{takahashi2020automatic} & 60.2 & 43.0 & 58.1 & 41.0 & 60.2 & 60.8 & 60.1 & 58.8 & 47.0 & 45.1 & 40.9 & \colorbox{blue!10}{52.3} \\
            
            \cdashline{1-13}\noalign{\vskip 0.1ex}
            
            \cmodel-UP & 60.1 & 44.4 & 50.7 & \textbf{45.3} & 62.2 & 62.1 & 61.1 & \textbf{61.5} & 48.3 & \textbf{47.3} & 42.3 & \colorbox{blue!10}{53.2} \\
            \cmodel-MUP & \textbf{62.1} & \textbf{45.6} & 52.2 & 44.8 & \textbf{62.5} & \textbf{63.0} & \textbf{61.9} & 61.4 & \textbf{49.1} & 46.9 & 42.3 & \colorbox{blue!10}{\textbf{53.8}} \\
            
            \midrule
            
            \multicolumn{13}{c}{\textit{Source-only Evaluation}} \\
            \cdashline{1-13}\noalign{\vskip 0.1ex}
            
            $^\heartsuit$MTransQuest~\cite{ranasinghe2020transquest} & 35.8 & 28.4 & 31.1 & 29.0 & 50.8 & 52.7 & 56.3 & 43.9 & 35.7 & 23.7 & ~~9.4 & \colorbox{red!10}{36.1} \\
            $^\diamondsuit$MTransQuest~\cite{ranasinghe2020transquest} & 40.2 & 33.1 & \textbf{32.9} & 32.8 & 54.2 & 57.2 & 60.2 & 49.1 & 40.4 & 29.8 & 17.9 & \colorbox{red!10}{40.7} \\
            $^\diamondsuit$XLM-R+Concat~\cite{takahashi2020automatic} & 53.5 & 38.0 & 30.2 & 34.0 & 53.9 & 55.9 & 53.5 & 53.8 & 35.7 & 32.5 & 31.5 & \colorbox{red!10}{42.9} \\
            
            \cdashline{1-13}\noalign{\vskip 0.1ex}
            
            \cmodel-UP & 52.3 & 41.7 & 27.3 & \textbf{40.7} & 60.7 & 59.1 & 60.4 & 56.8 & 40.7 & \textbf{37.0} & 32.1 & \colorbox{red!10}{46.3} \\
            \cmodel-MUP & \textbf{55.9} & \textbf{43.8} & 28.7 & 40.6 & \textbf{61.9} & \textbf{60.5} & \textbf{61.1} & \textbf{59.3} & \textbf{41.4} & 35.6 & \textbf{36.7} & \colorbox{red!10}{\textbf{47.8}} \\
            
            \midrule
            
            \multicolumn{13}{c}{\textit{Source-Reference-Combined Evaluation}} \\
            \cdashline{1-13}\noalign{\vskip 0.1ex}
            
            $^\diamondsuit$XLM-R+Concat~\cite{takahashi2020automatic} & 60.9 & 43.3 & 53.3 & 40.8 & 60.4 & 60.1 & 59.1 & 59.3 & 46.4 & 44.9 & 40.5 & \colorbox{green!10}{51.7} \\
            $^\diamondsuit$COMET~\cite{rei2020comet} & 61.0 & 44.6 & \textbf{58.3} & 42.3 & 62.3 & 60.7 & 59.0 & 60.6 & 45.7 & 46.8 & 38.8 & \colorbox{green!10}{52.7} \\
            
            \cdashline{1-13}\noalign{\vskip 0.1ex}
            
            \cmodel-UP & 60.0 & 44.9 & 49.7 & \textbf{45.6} & 62.7 & 62.6 & 62.0 & 61.0 & 48.0 & 45.5 & \textbf{42.4} & \colorbox{green!10}{53.1} \\
            \cmodel-MUP & \textbf{62.2} & \textbf{46.0} & 54.6 & 44.9 & \textbf{63.2} & \textbf{63.2} & \textbf{63.0} & \textbf{61.8} & \textbf{48.7} & \textbf{47.5} & \textbf{42.4} & \colorbox{green!10}{\textbf{54.3}} \\

    
            \bottomrule
        \end{tabular}
    }
    \caption{Kendall's Tau correlation (\%) on multilingual-targeted language pairs from WMT 2019 Metrics Shared Task. Even only pretrained with English-targeted data, \cmodel-UP model can still give competitive performance on multilingual-targeted tasks. \cmodel-MUP, which adopts multilingual-targeted pretraining, achieves new state-of-the-art results. Furthermore, \cmodel-UP and \cmodel-MUP are both single models, offering considerate convenience against other task-specific baselines.
    }
    \label{table.main_results_metric_en-x}
\end{table*}

        
        


        
\section{Experiments}
\subsection{Experimental Settings}
\paragraph{Benchmarks} Following~\newcite{rei2020comet, yuan2021bartscore}, we examine the effectiveness of the propose method on WMT 2019 Metrics~\cite{ma2019results}.
For the former, we follow the common practice in COMET\footnote{\href{https://github.com/Unbabel/COMET}{https://github.com/Unbabel/COMET}}~\cite{rei2020comet} to collect and preprocess the dataset.
The official variant of Kendall's Tau correlation~\cite{ma2019results} is used for evaluation.
We evaluate our methods on all of \cref, \csrc~and \csrcref~scenarios.
For \csrc~scenario, we further conduct results on WMT 2020 QE task~\cite{specia2020findings} referring to~\newcite{ranasinghe2020intelligent} for data collection and preprocessing.
Following the official report, the Pearson's correlation is used for evaluation.

\paragraph{Model Pretraining}
As mentioned in \S\ref{sec:up}, we continuously pretrain PLMs using synthetic data.
The data is constructed from WMT 2021 News Translation task, where we collect the training sets from five translation tasks.
Among those tasks, the target sentences are all in English (En), and the source languages are Czech (Cs), German (De), Japanese (Ja), Russian (Ru), and Chinese (Zh).
Specifically, we follow~\newcite{sellam2020bleurt} to use \textsc{Transformer}-base~\cite{vaswani2017attention} MT models to generate translation candidates, and use the checkpoints trained via \cmodel-MRA approach for synthetic data labeling.
We pretrain two kinds of models, one is pretrained on English-targeted language directions, and the other is a multilingual version trained using bidirectional data.
Note that, for a fair comparison, we filter out all pretraining examples that are involved in benchmarks.

\paragraph{Model Setting}
We implement our approach upon COMET~\cite{rei2020comet} repository and follow their work to choose XLM-R~\cite{conneau2020unsupervised} as the PLM.
The feedforward network consists of 3 linear transitions, where the dimensionalities of corresponding outputs are 3,072, 1,024, and 1, respectively.
Between any two adjacent linear modules inside, hyperbolic tangent function is arranged as activation.
During both pretraining and finetuning phrases, we divided training examples into three sets, where each set only serves one scenario among \cref, \csrc~and \csrcref~to avoid learning degeneration.
During finetuning, we randomly extracting 2,000 training examples from benchmarks as development set.
Besides \cmodel-MRA and \cmodel-UP which are derived with MRA (\S~\ref{sec:mra}) and Unified Pretraining (\S~\ref{sec:up}), we also extend the latter with multilingual-targeted unified pretraining, thus obtaining \cmodel-MUP model.

\paragraph{Baselines}
As to \cref~approaches, we select BLEU~\cite{papineni2002bleu}, ChrF~\cite{popovic2015chrf}, YiSi-1~\cite{lo2019yisi}, BERTScore~\cite{zhang2020bertscore}, BLEURT~\cite{sellam2020bleurt}, PRISM-ref~\cite{thompson2020automatic}, BARTScore~\cite{yuan2021bartscore}, XLM-R+Concat~\cite{takahashi2020automatic}, and RoBERTa+Concat~\cite{takahashi2020automatic} for comparison.
For \csrc~methods, we post results of both metric and QE methods, including YiSi-2~\cite{lo2019yisi}, XLM-R+Concat~\cite{takahashi2020automatic}, PRISM-src~\cite{thompson2020automatic} and multilingual-to-multilingual MTransQuest~\cite{ranasinghe2020transquest}.
For \csrcref, we use XLM-R+Concat~\cite{takahashi2020automatic} and COMET~\cite{rei2020comet} as strong baselines.


\subsection{Main Results}
\paragraph{English-Targeted} Results on English-targeted metric task are conducted in Table~\ref{table.main_results_metric_x-en}.
Among all involved baselines, for \cref~methods, BARTScore~\cite{yuan2021bartscore} performs better than other statistical and model-based metrics.
As to \csrc~scenario, MTransQuest~\cite{ranasinghe2020transquest} gives dominant performance.
Further, COMET~\cite{rei2020comet} performs better than XLM-R+Concat~\cite{takahashi2020automatic} on \csrcref~scenario.

As for our methods, we can see that, \cmodel-MRA achieves better results on all tasks, demonstrating the effectiveness of monotonic attention flows for cross-lingual interactions.
Moreover, the proposed model \cmodel-UP, which unifies \cref, \csrc, and \csrcref~learning on both pretraining and finetuning, yields better results on all evaluation settings.
Most importantly, \cmodel-UP is a \textit{single model} which surpasses all the different state-of-the-art models on three tasks, showing its dominance on both convenience and effectiveness.

\begin{table*}[t]
    \small
    \centering
    \begin{tabular}{lcccccccc}
        \toprule

        
        \textbf{Model} & \textit{\underline{De-En}} & \textit{\underline{Ru-En}} & \textit{\underline{Zh-En}} & \underline{Fi-En} & Gu-En & Kk-En & Lt-En & \textbf{Avg.} \\
        
        \midrule
        
        \multicolumn{9}{c}{\textit{Reference-only Evaluation}} \\
        \cdashline{1-9}\noalign{\vskip 0.1ex}
        \cmodel-MUP & 25.5 & 21.3 & 46.6 & 37.0 & \textbf{32.2} & \textbf{39.1} & 38.6 & \colorbox{blue!10}{34.3} \\
        \cmodel-UP & \textbf{25.6} & \textbf{21.9} & \textbf{46.7} & \textbf{37.9} & \textbf{32.2} & 38.7 & \textbf{40.0} & \colorbox{blue!10}{\textbf{34.8}} \\
        
        \midrule
        
        \multicolumn{9}{c}{\textit{Source-only Evaluation}} \\
        \cdashline{1-9}\noalign{\vskip 0.1ex}
        \cmodel-MUP & 18.0 & 16.3 & 41.0 & 33.9 & 29.6 & \textbf{34.7} & \textbf{35.7} & \colorbox{red!10}{29.9} \\
        \cmodel-UP & \textbf{19.3} & \textbf{16.9} & \textbf{41.4} & \textbf{34.0} & \textbf{29.7} & 33.6 & 35.4 & \colorbox{red!10}{\textbf{30.0}} \\
        
        \midrule
        
        \multicolumn{9}{c}{\textit{Source-Reference-Combined Evaluation}} \\
        \cdashline{1-9}\noalign{\vskip 0.1ex}
        \cmodel-MUP & 25.2 & 20.9 & 46.9 & 37.0 & 32.0 & 38.5 & 38.8 & \colorbox{green!10}{34.2} \\
        \cmodel-UP & \textbf{26.0} & \textbf{22.0} & \textbf{47.2} & \textbf{37.7} & \textbf{32.3} & \textbf{39.4} & \textbf{40.0} & \colorbox{green!10}{\textbf{35.0}} \\
        \bottomrule
    \end{tabular}
    \caption{Kendall's Tau correlation (\%) of semantic evaluation methods over English-targeted language pairs from WMT'19 Metrics Task test set. Compared to \cmodel-UP, \cmodel-MUP shows performance decrease over all translation tasks, yet still outperforms all related baselines in Table~\ref{table.main_results_metric_x-en}.}
    \label{table.appendix_model-m_english-targeted}
\end{table*}
\paragraph{Multilingual-Targeted}
As seen in Table~\ref{table.main_results_metric_en-x}, the multilingual-targeted \cmodel-MUP gives dominant performance than all strong baselines on \cref, \csrc~and \csrcref, demonstrating the transferability and effectiveness of our approach.
Besides, the \cmodel-UP also gives dominant results, revealing an improvement of 0.6, 0.3 and 0.9 averaged Kendall's $\tau$ correlation scores, respectively.
However, we find that \cmodel-MUP outperforms strong baselines but slightly worse than \cmodel-UP on English-targeted translation directions (see Table~\ref{table.appendix_model-m_english-targeted}).
We think the reason lies in the curse of multilingualism and vocabulary dilution~\cite{conneau2020unsupervised}.

\begin{table*}[t]
    \small
    \centering
    \begin{tabular}{lcccccccc}
        \toprule
        \textbf{Model} & \textit{{En-De}} & \textit{En-Zh} & \textit{{Ru-En}} & Et-En & Ne-En & Ro-En & Si-En & \textbf{Avg.} \\
        \midrule
        \noalign{\vskip -0.5ex}
        OpenKiwi~\cite{kepler2019openkiwi} & 14.6 & 19.0 & 54.8 & 47.7 & 38.6 & 68.5 & 37.4 & \colorbox{red!10}{40.1} \\
        mBERT~\cite{devlin2019bert} & 37.7 & 39.8 & 66.6 & 62.3 & 64.5 & 83.5 & - & \colorbox{red!10}{-} \\
        \noalign{\vskip -0.5ex}
        TransQuest-m~\cite{ranasinghe2020transquest} & 44.2 & 46.5 & \textbf{75.2} & 75.7 & \textbf{75.8} & \textbf{88.6} & \textbf{65.3} & \colorbox{red!10}{67.3} \\
        \cdashline{1-9}\noalign{\vskip 0.1ex}
        \cmodel-MUP & \textbf{52.5} & \textbf{50.5} & 64.4 & \textbf{79.1} & 75.6 & 88.3 & 64.3 & \colorbox{red!10}{\textbf{67.8}} \\
        
        \bottomrule
    \end{tabular}
    \caption{Pearson correlation (\%) on WMT 2020 QE Task test set. For baselines, we directly collect the results reported in~\newcite{ranasinghe2020transquest}. \cmodel-MUP gives better results between convenience and performance.}
    \label{table.main_results_qe}
\end{table*}
\paragraph{Quality Estimation}
The results for \cmodel~approach on WMT 2020 QE task are concluded in Table~\ref{table.main_results_qe}.
As seen, it achieves competitive results on QE task compared with the winner submission~\cite{ranasinghe2020transquest}.

\section{Ablation Studies}
In this section, we conduct ablation studies to investigate the effectiveness of regional attention patterns (\S\ref{sec.ablation.mra}), unified training (\S\ref{sec:unift}), and ranking-based data labeling (\S\ref{sec:rdl}).
All experiments are conducted by following English-targeted setting.

\subsection{Regional Attention Patterns}
\label{sec.ablation.mra}

\begin{table}[t]
    \centering
    \small
    \scalebox{1.0}
    {
        \begin{tabular}{lcr}
            \toprule
            \textbf{Model} & \textbf{Avg. $\tau$} (\%) & $\Delta$ \\
            \midrule
            Full attention & 34.1 & -- \\
            \cdashline{1-3}\noalign{\vskip 0.3ex}
            ~~no H$\rightarrow$S (Soft) & \textbf{34.8} & \textbf{+0.7} \\
            ~~no S$\rightarrow$H (Soft) & 34.6 & +0.5 \\
            \cdashline{1-3}\noalign{\vskip 0.3ex}
            ~~no H$\rightarrow$S, H$\rightarrow$R \& S$\rightarrow$R (Hard) & 34.3 & +0.2 \\
            \cdashline{1-3}\noalign{\vskip 0.3ex}
            ~~no R$\rightarrow$S & 34.0 & -0.1 \\
            ~~no S$\rightarrow$R & 33.9 & -0.2 \\
            ~~no R$\rightarrow$H & 33.6 & -0.5 \\
            ~~no H$\rightarrow$R & 34.0 & -0.1 \\
            \bottomrule
        \end{tabular}
    }
    \caption{Averaged Kendall's Tau correlation (\%) and the gap ($\Delta$) on English-targeted \fboxsep0.5ex\csrcref~task with monotonic regional attention (MRA) strategies. H, S and R represent hypothesis, source and reference segment, respectively. Arrow denotes the attention flow of two segments inside attention modules of XLM-R. Soft MRA strategy between H and S is most effective. Hard MRA can yield a slight improvement. Removing other interactions between H and R, or S and R, leads to performance drop, and R$\rightarrow$H degrades most.}
    \label{table.ablation_monotonic}
\end{table}
To investigate the effectiveness of MRA, we further collect experiments in Table~\ref{table.ablation_monotonic}.
As seen, MRA can give performance improvements than full attention, and preventing the interactions between hypothesis and source segment can improve the performance most.
We think the reasons behind are twofold.
First, the source side is formed with a different language, whose semantic information is rather weak than the reference side.
Second, by preventing direct interactions between source and hypothesis, semantics inside the former must be passed through reference, which is helpful for disambiguation.
Besides, not allowing the source to derive information from the hypothesis is better than the opposite direction.
\newcite{wang2020what} found that the positional embeddings in PLM are engaged with strong adjacent information.
We think the reason why S$\rightarrow$H performs worse than H$\rightarrow$S lies in the skipping of indexes, which corrupts positional similarities in alignment calculation.

Additionally, when we combined two methods together, \textit{i.e.}, unified pretraining and finetuning with \csrcref~\cmodel-MRA setting, model performance drops to 34.9 over English-targeted tasks on average.
We think that both methods all intend to solve the problem of unseen \csrcref~input format, and MRA may not be necessary if massive data examples can be obtained for pretraining.
Nevertheless, \cmodel-MRA has its advantage on wide application without requiring pseduo labeled data.

\subsection{Unified Training}
\label{sec:unift}
        
        
        

\begin{table}[t]
    \centering
    \small
    \scalebox{1.0}{
        \begin{tabular}{cccccr}
            \toprule
            \textbf{Unified} & \textbf{Unified} & \multicolumn{3}{c}{\textbf{Avg. $\tau$} (\%)} \\
            \textbf{Pretrain} & \textbf{Finetune} & \ccref & \ccsrc & \ccsrcref \\
            \midrule
            
            \ding{51} & \ding{51} & \textbf{34.8} & \textbf{30.0} & \textbf{35.0} \\
            
            \cdashline{1-5}\noalign{\vskip 0.3ex}
            \ding{51} & \ding{55} & 33.8 & 29.1 & 33.9 \\
            
            
            \ding{55} & \ding{55} & 31.9 & 27.7 & 32.6 \\
            
            \bottomrule
        \end{tabular}
    }
    \caption{Unified and task-specific training for \cmodel-UP approach. As seen, combination of unified pretraining and finetuning gives best performances, meanwhile requires only one unified model.}
    \label{table.ablation_unified_specific_training}
\end{table}
\begin{table}[t]
    \centering
    \small
    \scalebox{1.0}
    {
        \begin{tabular}{lcr}
            \toprule
            \textbf{Method} & \textbf{Avg.} $\tau$ (\%)  & $\Delta$ \\
            \midrule
            Rank-Norm, Ens & \textbf{35.0} & - \\
            Rank-Norm & 34.7 & -0.3 \\
            \cdashline{1-3}\noalign{\vskip 0.3ex}
            Z-Norm, Ens & 33.5 & -1.5 \\
            Z-Norm & 34.2 & -0.8 \\
            
            \bottomrule
        \end{tabular}
    }
    \caption{Pseudo-data labeling with different methods. Ranking-based normalization (Rank-Norm) performs better than conventional Z-score approach (Z-Norm). Besides, ensembling (Ens) ranking-based normalized scores can give higher result, while conventional Z-Norm performs worse after ensembling.}
    \label{table.ablation_pseudo_labeling}
\end{table}

Experiments for comparing unified and task-specific training are concluded in Table~\ref{table.ablation_unified_specific_training}.
As seen, when using the unified pretraining checkpoint to finetune over the specific task, performance over three models reveals performance drop consistently, indicating that the unified finetuning is helpful for model learning.
This also verifies our hypothesis, that the cores of \cref, \csrc, and \csrcref~tasks are identical to each other.
Moreover, unified pretraining and finetuning are complementary to each other.
Also, utilizing task-specific pretraining instead of unified one reveals worse performance.
To sum up, unifying both pretraining and finetuning only reveals one model, showing its advantage on the generalization on all tasks, where one united model can cover all functionalities of \cref, \csrc~and \csrcref~tasks concurrently.

\subsection{Ranking-based Data Labeling}
\label{sec:rdl}
To verify the effectiveness of ranking-based labeling, we collect the results of models applying different pseudo labeling strategies.
After deriving the original scores from the well-trained \cmodel-MRA checkpoint, we use Z-score and proposed ranking-based normalization methods to label synthetic data.
For both methods, we also apply an ensembling strategy to assign training examples with averaged scores deriving from 3 \cmodel-MRA checkpoints.
Results show that, Z-score normalization reveals a performance drop when applying score ensembling with multiple models.
Our proposed ranking-based normalization can boost the \cmodel-UP model training, and its ensembling approach can further improve the performance.

\section{Conclusion}
In the past decades, automatic translation evaluation is mainly divided into \cref, \csrc~and \csrcref~tasks, each of which  develops independently and is tackled by various task-specific methods.
We suggest that the three tasks are possibly handled by a unified framework, thus being ease of use and facilitating the knowledge transferring.
Contributions of our work are mainly in three folds:
(a) We propose a flexible and unified translation evaluation model UniTE, which can be adopted into the three tasks at once;
(b) Through in-depth analyses, we point out that the main challenge of unifying three tasks stems from the discrepancy between vanilla pretraining and multi-tasks finetuning, and fill this gap via monotonic regional attention (MRA) and unified pretraining (UP);
(c) Our single model consistently outperforms a variety of state-of-the-art or winner systems across high-resource and zero-shot evaluation in WMT 2019 Metrics and WMT 2020 QE benchmarks, showing its advantage of flexibility and convincingness.
We hope our new insights can contribute to subsequent studies in the translation evaluation community.

\section*{Acknowledgements}
The authors would like to send great thanks to all reviewers and meta-reviewer for their insightful comments.
This work was supported in part by the Science and Technology Development Fund, Macau SAR (Grant No. 0101/2019/A2), the Multi-year Research Grant from the University of Macau (Grant No. MYRG2020-00054-FST), National Key Research and Development Program of China (No. 2018YFB1403202), and Alibaba Group through Alibaba Research Intern Program.

\bibliographystyle{acl_natbib}
\bibliography{anthology}

\appendix

\section{Collection of Pretraining Data} 
Considering the English-targeted model, we select Czech (Cz), German (De), Japanese (Ja), Russian (Ru), and Chinese (Zh) as source languages, and English (En) as target.
For each translation direction, we collect 1 million samples, finally yielding 5 million examples in total for unified pretraining. 
As to the multilingual-targeted model, we further  collect 1 million synthetic data for each language direction of En-Cz, En-De, En-Ja, En-Ru, and En-Zh. Finally, we construct 10 million examples for the pretraining of the multilingual version by adding the data of the English-targeted model.
Note that, for a fair comparison, we filter out all pretraining examples that are involved in benchmarks.




\section{Reproducibility}
All the models reported in this paper were finetuned on a single Nvidia V100 (32GB) GPU.
Specifically for \cmodel-UP and \cmodel-MUP, the pretraining is arranged on 4 Nvidia V100 (32GB) GPUs. 
Our framework is built upon COMET repository~\cite{rei2020comet}.
For the contribution to the research community, we release both the source code of \cmodel~framework and the well-trained evaluation models as described in this paper at \href{https://github.com/NLP2CT/UniTE}{https://github.com/NLP2CT/UniTE}.

\end{document}